\documentclass[journal]{IEEEtran}

\usepackage{amsmath, amsfonts, amssymb} 
\usepackage{algorithmic}
\usepackage{array}
\usepackage{subcaption} 
\usepackage{textcomp}
\usepackage{stfloats}
\usepackage{url}
\usepackage{verbatim}
\usepackage{graphicx}
\usepackage{cite}
\usepackage{multirow}
\usepackage{booktabs}
\usepackage{bm}
\usepackage[ruled]{algorithm2e}
\usepackage{float}
\usepackage{indentfirst}
\usepackage{color}
\usepackage{makecell}
\usepackage{adjustbox}

\begin{document}

\title{Stereo Event-based, 6-DOF Pose Tracking for Uncooperative Spacecraft}

\author{Zibin Liu, Banglei Guan,~\IEEEmembership{Member,~IEEE}, Yang Shang, Yifei Bian, Pengju Sun and Qifeng Yu
	\thanks{This work was supported in part by the National Natural Science Foundation of China under Grant 12372189, and the Hunan Provincial Natural Science Foundation for Excellent Young Scholars under Grant 2023JJ20045. \textit{(Corresponding authors: Banglei Guan; Yang Shang)}

	Zibin Liu, Banglei Guan, Yang Shang, Yifei Bian, Pengju Sun and Qifeng Yu are with the College of Aerospace Science and Engineering, National University of Defense Technology, Changsha 410073, China (e-mail: liuzibin@nudt.edu.cn; guanbanglei12@nudt.edu.cn; shangyang1977@nudt.edu.cn; bianyifei18@nudt.edu.cn; pengjusun@yeah.net; yuqifeng@nudt.edu.cn).}
}

\maketitle

\begin{abstract}
Pose tracking of uncooperative spacecraft is an essential technology for space exploration and on-orbit servicing, which remains an open problem. Event cameras possess numerous advantages, such as high dynamic range, high temporal resolution, and low power consumption. These attributes hold the promise of overcoming challenges encountered by conventional cameras, including motion blur and extreme illumination, among others. To address the standard on‐orbit observation missions, we propose a line-based pose tracking method for uncooperative spacecraft utilizing a stereo event camera. To begin with, we estimate the wireframe model of uncooperative spacecraft, leveraging the spatio-temporal consistency of stereo event streams for line-based reconstruction. Then, we develop an effective strategy to establish correspondences between events and projected lines of uncooperative spacecraft. Using these correspondences, we formulate the pose tracking as a continuous optimization process over 6-DOF motion parameters, achieved by minimizing event-line distances. Moreover, we construct a stereo event-based uncooperative spacecraft motion dataset, encompassing both simulated and real events. The proposed method is quantitatively evaluated through experiments conducted on our self-collected dataset, demonstrating an improvement in terms of effectiveness and accuracy over competing methods. The code will be open-sourced at \url{https://github.com/Zibin6/SE6PT}.
\end{abstract}

\begin{IEEEkeywords}
Uncooperative spacecraft, pose tracking, stereo event camera.
\end{IEEEkeywords}

\section{Introduction}
\label{sec1}

\begin{figure*}[t]
	\centering
	\includegraphics[width=18cm,height=7cm]{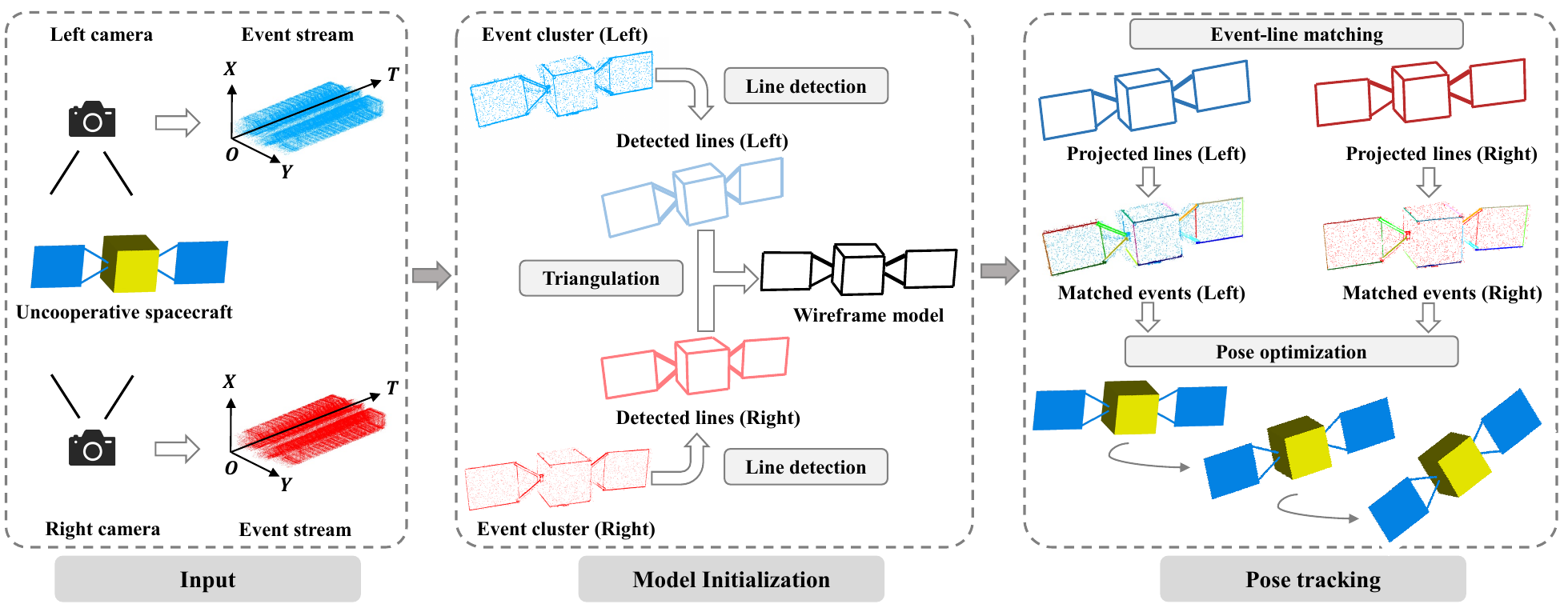}
	\caption{Block diagram of our method. The core of our method consists of model initialization and pose tracking. Model initialization is a one-time operation, followed by pose tracking. Our method takes stereo event streams and stereo camera parameters as input and provides the pose and trajectory of uncooperative spacecraft as output.}
	\label{fig1}
\end{figure*}

Over the past few years, there has been a noticeable increase in the frequency of space exploration activities and the number of space launch missions, leading to a significant rise in the quantity of uncooperative spacecraft in orbit. Uncooperative spacecraft are entities that reside in space, without establishing any form of communication or collaborative relationships with other spacecraft or space systems~\cite{opromolla2017review,Zhan2024TimetogoBT}. These entities include space debris, unauthorized spacecraft, damaged and abandoned satellites, among others. Pose tracking of uncooperative spacecraft is a critical task for a wide range of space missions, such as active debris removal~\cite{6809997}, on-orbit servicing~\cite{8864103,8486983,10234379}, rendezvous and docking~\cite{fehse2003automated}. Owing to the lack of priori structures and cooperative marks, the pose tracking of uncooperative spacecraft is still an open challenge~\cite{cassinis2019review}.

Electro-Optical sensors, such as monocular~\cite{pauly2023survey} and binocular cameras~\cite{segal2013stereovision}, as well as lidars~\cite{8944033,9733900}, are widely regarded as the best choice for pose determination of spacecraft in close-proximity~\cite{opromolla2017review}. However, traditional sensors still encounter numerous challenges when dealing with uncooperative targets. These challenges include, but are not limited to: motion blur, particularly when a spacecraft undergoes high-speed or uncontrollable motion~\cite{cassinis2019review}; extreme lighting, like strong light sources from the sun and other celestial bodies, introducing interference into the observation process~\cite{liu2021three,9340601}. This prompts us to ponder whether there exist more effective visual solutions to address these challenges.

A new type of bio-inspired vision sensor, called ``event cameras", has emerged as a potential solution to the challenges outlined above. Unlike standard cameras that capture images at unnatural fixed time periods, event cameras asynchronously trigger events, when the brightness change of an individual pixel exceeds a predefined threshold. The unique imaging principle endows event cameras with numerous advantages over standard cameras: high temporal resolution, high dynamic range, low latency and low power consumption~\cite{gallego_event-based_2022}. Event cameras demonstrate remarkable advantages in space observation, presenting significant potential for event-based visual applications~\cite{huang20221000,yu2024robust}, particularly in star tracking~\cite{Chin_2019_CVPR_Workshops} and satellite pose estimation~\cite{jawaid2022towards}. When dealing with the pose tracking of uncooperative spacecraft, it becomes vital to address the following key issues.

The primary issue is the absence of a priori knowledge regarding the 3D structure of uncooperative spacecraft, making it difficult to directly determine the pose of spacecraft through 2D-3D structure correspondences. Due to the inherent depth uncertainty in monocular setups, it is difficult to directly determine the object pose with absolute scale relying solely on a single camera~\cite{Guan_TCYB2021,guan2022minimal_IJCV,guan2022monitoring}. As a result, we employ a stereo event camera rig as our input modality, allowing for 6 Degrees-Of-Freedom (DOF) pose estimation and continuous tracking of uncooperative spacecraft.
\IEEEpubidadjcol

Another key issue is the lack of cooperative marks and known texture patterns on uncooperative spacecraft. This presents a significant obstacle to the stable extraction and continuous tracking of event-based feature points. On the other hand, uncooperative spacecraft usually contain a substantial quantity of lines depicting their geometric structures. This prompts consideration of leveraging the lines of uncooperative spacecraft for pose tracking. Lines demonstrate higher resistance to variations in illumination conditions, and are less susceptible to interference from clutter and noise~\cite{opromolla2017review}. Furthermore, event cameras are predominantly sensitive to lines and edges of objects or scenes~\cite{liu2024line}. These intrinsic properties make lines an ideal feature for event-based pose tracking of uncooperative spacecraft.

In this paper, we propose a line-based pose tracking method for uncooperative spacecraft using a stereo event camera. This method is specifically designed for close-range observations using space-based observing systems. Firstly, we conduct an initialization process to obtain the 3D line model of uncooperative spacecraft. Subsequently, we establish associations between events and projected lines of the reconstructed model. Ultimately, we formulate the pose tracking as a continuous optimization process over 6-DOF motion parameters. Using bundle adjustment, the pose of uncooperative spacecraft can be refined by minimizing the distances between events and lines.
A block diagram of the proposed method with key steps is shown in Fig.~\ref{fig1}. 
Extensive experiments are demonstrated based on our self-collected datasets, which consist of simulated events from various spacecraft as well as real events captured using two Prophesee EVK4 event cameras, demonstrating the superiority of our method over the baselines. The main contributions are listed as follows:

\begin{itemize}
	\item To the best of our knowledge, this is the first pose tracking method for uncooperative spacecraft using a stereo event camera. It relies solely on event streams without any intermediate products (e.g., reconstructed images), resulting in a simplified workflow.
	\item We propose an event-based model initialization method to reconstruct the wireframe model of uncooperative spacecraft, leveraging the spatio-temporal consistency of stereo event streams.  
	\item We present a stereo pose optimization method by establishing event-line matching and minimizing event-line distances. To evaluate our method, we build the first stereo event dataset for uncooperative spacecraft, derived from various types of spacecraft with diverse trajectories.
\end{itemize}

\section{Related Work}
\label{sec2}
In recent years, there have been numerous event-based solutions addressing various aspects. We revisit classical event-based tracking methods, including solutions for event-based feature tracking, object tracking, and camera tracking.

\textbf{Event-Based Feature Tracking.}
In order to fully harness the potential of event cameras and address various challenges in computer vision, event-based feature tracking has been extensively studied over the past years, such as corners~\cite{7759610}, circles~\cite{huang_dynamic_2021}, balls~\cite{7759345} and lines~\cite{elised_2016}. Advancements have been made in extending widely-used image-based keypoint detectors to accommodate the distinctive input modalities of event cameras, such as luvHarris~\cite{glover2021luvharris}, FA-Harris~\cite{li2019fa} and eFAST~\cite{mueggler2017fast}. Moreover, Huang et al.~\cite{huang_dynamic_2021} perform circle detection for event camera calibration, relying on noise-resilient cluster detection as well as circle fitting. Furthermore, Hough transform~\cite{conradt2009pencilf} has already been utilized for event-based line detection. Conradt et al.~\cite{conradt2009pencilf} represent the detected line as a Gaussian within the Hough space, and subsequently update these lines by incoming events. Event-Based Line Segment Detector (ELiSeD)~\cite{elised_2016}, derived from the LSD algorithm~\cite{von2012lsd}, detects and tracks line segments in an event-by-event fashion. Their method demonstrates that line features are distinct enough to be utilized as trackers even in dynamic scenes.

\textbf{Event-Based Object Tracking.}
In the early work, a stationary camera setup is utilized to capture the motion of planar objects, followed by pose tracking using shape templates, either pre-defined~\cite{ni2012asynchronous} or built from events~\cite{zhu2017event}. Iterative methods are adopted to update the object pose by associating new events with the object shape, employing a nearest-neighbor strategy~\cite{ni2015visual}. Recently, Valeiras et al.~\cite{reverter2016neuromorphic} propose a 6-DoF pose estimation method for rigid objects. Their method achieves continuous object tracking by aligning events with the closest edges of the object, given its 3D wireframe model and initial pose. Kang et al.~\cite{Yufan2024} propose the object tracking method by direct brightness increment alignment with motion interpolation. Glover et al.~\cite{glover2024} introduce the EDOPT method, leveraging the Exponentially Reduced Ordinal Surface (EROS) of events, capable of real-time object pose tracking. Without relying on a fixed number of events or a predetermined time interval, EROS can more effectively maintain the persistence of object edges over time. In the domain of space exploration, the feasibility of event-based star tracking~\cite{Chin_2019_CVPR_Workshops} has been confirmed, while the application of event cameras for spacecraft pose estimation remains at an early stage of development. Jawaid et al.~\cite{jawaid2022towards} propose an event-based satellite pose estimation method, using classic learning-based approaches for feature detection, followed by utilizing the PnP solver for pose estimation. Prior knowledge of the satellite structure is required for these methods. Furthermore, there are event-based datasets available for tracking object motion, including common objects~\cite{glover2024} and satellites~\cite{Arunkum2024}.

As far as we know, there are currently no existing methods for object pose tracking with stereo event cameras.

\textbf{Event-Based Camera Tracking.}
Early event-based camera tracking focuses more on simple 3-DOF motion estimation (e.g., planar~\cite{weikersdorfer2013simultaneous} or rotational~\cite{bryner2019event}). Recently, there have been attempts to tackle camera tracking in 6-DOF. Mueggler et al.~\cite{mueggler2014event} track the pose of a Dynamic Vision Sensor (DVS) during high-speed motion. Their method heavily relies on artificial line-based maps and uses the corners of intersecting lines for pose estimation. Subsequently, Bryner et al.~\cite{bryner2019event} demonstrate the ability to track 6-DOF high-speed motion in natural scenes. Their method utilizes the events within a maximum-likelihood framework to estimate the camera motion in a known environment, employing non-linear optimization to minimize the photometric error. The event-based stereo visual odometry system is first proposed by Zhou et al.~\cite{9386209}, which essentially employs a parallel tracking-and-mapping philosophy. The mapping module builds a semi-dense 3D scene map, and the tracking module determines the camera pose by addressing the 3D-2D registration problem. Building upon this, Niu et al.~\cite{junkai2024} integrate IMU data to present a direct visual-inertial odometry. A more compact event representation is introduced, called adaptive accumulation, which preserves relatively complete edges while maintaining a high signal-to-noise ratio.

Currently, the majority of event-based tracking methods primarily focus on aggregating events into images, thereby enabling the application or adaptation of traditional frame-based methods. To fully unlock the potential of this unique input modality, it is crucial to develop novel algorithms or paradigms that directly leverage event streams. Different from the aforementioned methods, we do not require prior knowledge of the object model. Besides that, our method works directly on events without intermediate quantities, thus fully utilizing the available information within events.

\section{Method}
\label{sec3}
Our method aims to continuously recover 6-DoF that defines the position and orientation of uncooperative spacecraft relative to the event camera or, equivalently, the camera pose relative to spacecraft. Initially, we perform the model initialization of uncooperative spacecraft to reconstruct its wireframe model in Section~\ref{sec3.1}. Subsequently, we employ the reconstructed model for pose tracking in Section~\ref{sec3.2}.

\subsection{Model initialization}
\label{sec3.1}

For uncooperative spacecraft where prior structural information is unknown, it is necessary to perform model initialization. Using calibrated stereo event cameras facilitates the reconstruction of the object's 3D structure. Firstly, we partition the stereo event stream and perform stereo event clustering. Subsequently, extract lines separately from the first event clusters of the left and right cameras. The initial wireframe model of uncooperative spacecraft can be obtained by line triangulation, followed by further optimization. The aforementioned process focuses on handling lines, culminating in endpoint determination to ascertain the endpoints of these lines.

\begin{figure}[t] 
	\centering{\includegraphics[width=8.5cm]{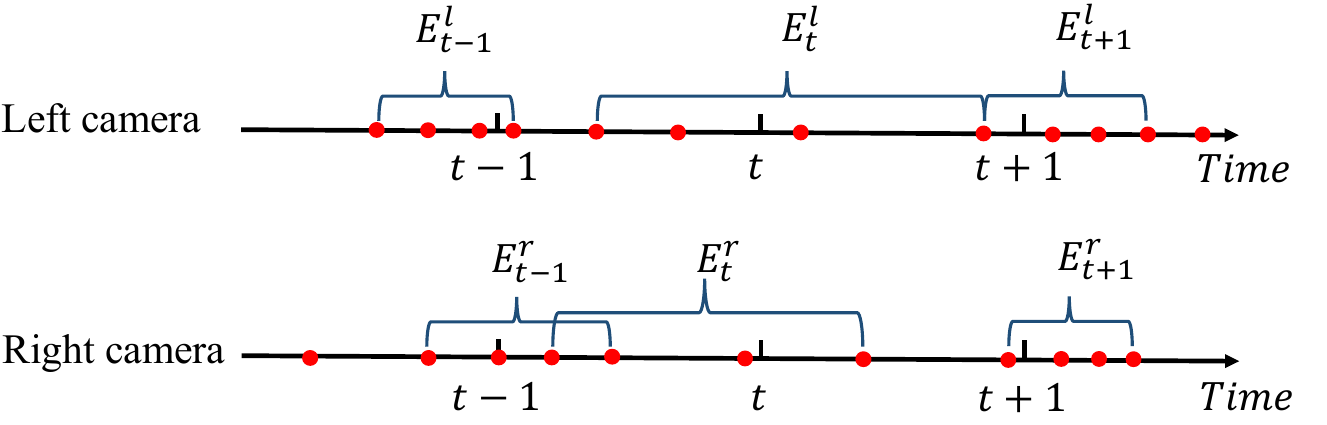}}
	\caption{The schematic diagram of stereo event clustering. The red dots represent individual events, while the blue brackets denote clusters of events. $\left\{ \mathbf{E}_{t-1}^{l}, \mathbf{E}_{t-1}^{r} \right\}, \left\{ \mathbf{E}_{t}^{l}, \mathbf{E}_{t}^{r} \right\}, \left\{ \mathbf{E}_{t+1}^{l}, \mathbf{E}_{t+1}^{r} \right\}$ represent event clusters for the left and right cameras at time $t-1$, $t$, and $t+1$, respectively.}
	\label{fig2}
\end{figure}

\textbf{Stereo event clustering.} 
Unlike traditional cameras, event cameras directly produce asynchronous event streams. Each event encodes the pixel coordinate, timestamp, and polarity of the brightness change. Events within a spatio-temporal window are generally collectively processed. Therefore, the selection of an appropriate window size is essential for event-based object tracking. Another aspect that needs to be considered is the association of stereo event streams. We solve these issues by creating two corresponding spatio-temporal windows of events $\mathbf{E}_{t}^{l}=\left\{ {\mathbf{e}_{j}^{l}} \right\}$ and $\mathbf{E}_{t}^{r}=\left\{ {\mathbf{e}_{k}^{r}} \right\}$ for the left and right cameras, respectively. The two windows consist of a fixed number of events that triggered in closest proximity to the time $t$. As illustrated in Fig.~\ref{fig2}, the red dots represent individual events, and the blue brackets denote clusters of events at different times. Our method effectively leverages a combined strategy of fixed time intervals between clusters and predetermined event quantities, thereby yielding a more robust representation of the object pose at time $t$. By using an event cluster instead of individual events, it becomes possible to mitigate the impact of stereo temporal disparities, thereby further ensuring the temporal consistency of stereo event streams. The first stereo event cluster is employed for model initialization, while subsequent clusters are utilized for pose tracking.

\textbf{Event-based line extraction.} Firstly, extract the lines of uncooperative spacecraft from the first stereo event cluster. During the brief temporal interval of the event cluster, events generated by lines within the space-time volume are approximately planar~\cite{everding_low-latency_2018}. Therefore, we consider events as a 3D point cloud for processing. The first event cluster is transformed into a 3D point cloud in the space-time volume, where the first two dimensions are the event’s pixel coordinates. The third dimension is represented by the event's trigger time, normalized by a constant ${c}_{z}$, and is expressed as ${t}/{c}_{z}$. This conversion effectively transforms the time dimension into a geometric one, resulting in a sparse representation. By tuning the constant ${c}_{z}$, we can effectively compress the point cloud along the time axis, achieving a more compact spatial-temporal representation. We employ an event based line extraction algorithm to detect lines from the event cluster~\cite{liu2024lecalib}. Lines are extracted from the first stereo event cluster, denoted as $\mathcal{L}_{l}$ and $\mathcal{L}_{r}$, respectively.

\begin{figure*}[tbp]
	\centering
	\subfloat[Line triangulation]{
		\begin{minipage}[t]{0.3\textwidth}
			\centering
			\includegraphics[width=\linewidth]{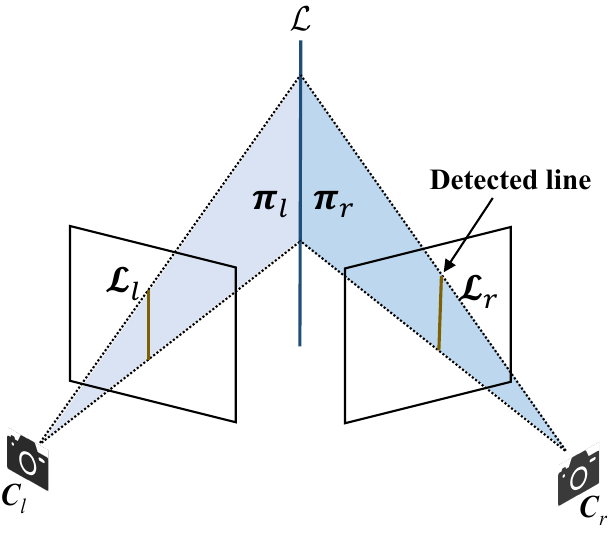}
			\label{fig3a}
		\end{minipage}%
	}
	\hfill
	\subfloat[Line optimization]{
		\begin{minipage}[t]{0.3\textwidth}
			\centering
			\includegraphics[width=\linewidth]{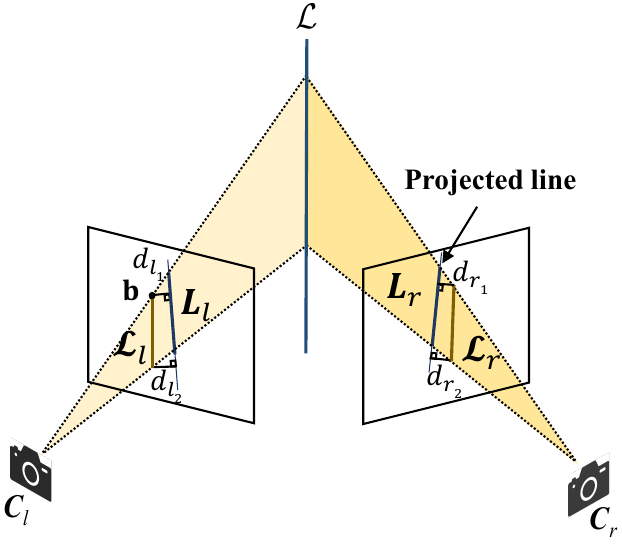}
			\label{fig3b}
		\end{minipage}%
	}
	\hfill
	\subfloat[Endpoint determination]{
		\begin{minipage}[t]{0.32\textwidth}
			\centering
			\includegraphics[width=\linewidth]{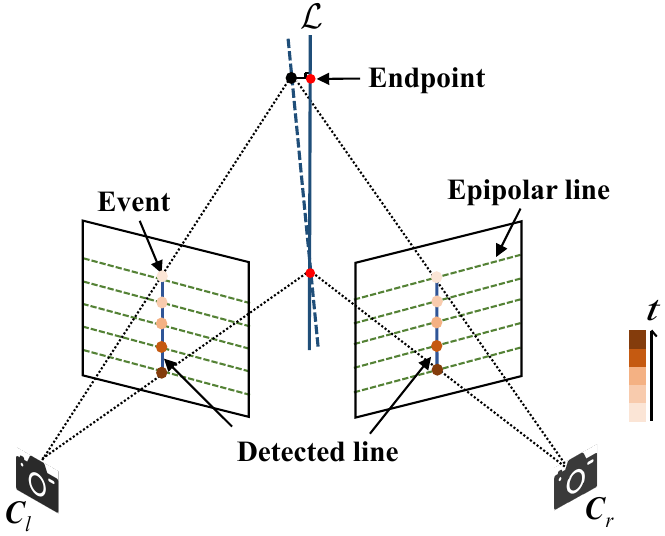}
			\label{fig3c}
		\end{minipage}%
	}
	\caption{The geometric interpretation of model initialization in a stereo camera rig. Initially, perform stereo event clustering, followed by the extraction of lines from the event cluster. Subsequently, proceed with the sequential execution of (a) line triangulation, (b) line optimization, and (c) endpoint determination.}
	\label{fig3}
\end{figure*}

\textbf{Line triangulation.}
Upon extracting lines from the event cluster, it is essential to establish line correspondences between the left and right cameras. Within confined spatial constraints, the typical configuration involves the installation of two cameras in close proximity, with their optical axes nearly parallel. In this stereo camera setup, we utilize the nearest-neighbor principle to search for the line that exhibits the closest proximity in both distance and slope, as well as midpoint, thus serving as the corresponding line from the other camera. Once the line correspondences have been successfully established, we proceed with line triangulation to derive the initial wireframe model of the uncooperative spacecraft.

Next, proceed to geometrically expound upon line triangulation. Given two pre-calibrated event cameras, their intrinsic matrices for points and lines are denoted as ${\mathbf{K}_{p}}$ and ${\mathbf{K}_{e}}$, respectively,
\begin{equation}
	{{\mathbf{K}}_{p}}=\left[ \begin{matrix}
		{{f}_{x}} & 0 & {{c}_{x}}  \\
		0 & {{f}_{y}} & {{c}_{y}}  \\
		0 & 0 & 1  \\
	\end{matrix} \right],{{\mathbf{K}}_{e}}=\left[ \begin{matrix}
		{{f}_{y}} & 0 & 0  \\
		0 & {{f}_{x}} & 0  \\
		-{{f}_{y}}{{c}_{x}} & -{{f}_{x}}{{c}_{y}} & {{f}_{x}}{{f}_{y}}  \\
	\end{matrix} \right].
	\label{eq1-0}
\end{equation}

These matrices include the focal lengths ${{f}_{x}}$, ${{f}_{y}}$, and principal points ${{\left( {{c}_{x}},{{c}_{y}} \right)}^{T}}$. The extrinsic parameters of the left and right cameras are expressed using the rotation matrix ${{\mathbf{R}}_{l}},{{\mathbf{R}}_{r}}$, and the translation vector ${{\mathbf{T}}_{l}},{{\mathbf{T}}_{r}}$, respectively. Let us denote the plane formed by a line $\mathcal{L}_{l}$ and left camera center $\mathbf{C}_{l}$ as ${{\bm{\pi }}_{l}}$, and the plane formed by the corresponding line $\mathcal{L}_{r}$ and the right camera center $\mathbf{C}_{r}$ as ${{\bm{\pi }}_{r}}$. As depicted in Fig.~\ref{fig3}\subref{fig3a}, the intersection of two planes yields a 3D line, which can be represented using the Plücker line coordinate as $\mathcal{L}={{\left( {{\mathbf{n}}^{T}},{{\mathbf{v}}^{T}} \right)}^{T}}$ in the world frame~\cite{hartley2003multiple}. In this representation, the vector $\mathbf{n}$ represents the normal vector to the plane that encompasses the line $\mathcal{L}$ and the origin, and $\mathbf{v}$ denotes the direction of the line $\mathcal{L}$. Given two back-projected planes ${{\bm{\pi }}_{l}}$ and ${{\bm{\pi }}_{r}}$, we can effectively compute the dual Plücker matrix ${{\mathcal{L}}^{*}}$ by

\begin{equation}
	\mathcal{L}^{*}={{\bm{\pi }}_{l}}{{\bm{\pi }}_{r}}^{T}-{{\bm{\pi }}_{r}}{{\bm{\pi }}_{l}}^{T}.
	%\in {{\mathbb{R}}^{4\times 4}}.
	\label{eq1-1}
\end{equation}

Based on the properties of the dual Plücker matrix

\begin{equation}
	\mathcal{L}^{*}=\left[ \begin{matrix}
		{{[\mathbf{v}]}_{\times }} & \mathbf{n}  \\
		-{{\mathbf{n}}^{T}} & 0  \\
	\end{matrix} \right],
	\label{eq1-2}
\end{equation}
where ${{[\cdot ]}_{\times}}$ is the antisymmetric form, the Plücker coordinates $\mathcal{L}={{\left( {{\mathbf{n}}^{T}},{{\mathbf{v}}^{T}} \right)}^{T}}$ can be recovered from Eq.~(\ref{eq1-2}).

\textbf{Line optimization.} The accuracy of line triangulation depends on the precision of line extraction. Due to noise interference from events, line extraction is not entirely precise, necessitating optimization of the reconstructed lines. Line optimization is conducted by minimizing the reprojection error of lines. For instance, given a tentative endpoint $\mathbf{b}={{\left[ {{x}_{b}},{{y}_{b}} \right]}^{T}}$ of extracted lines and its corresponding projected line $\mathbf{L}_l$, the error $d_{l_1}$ can be computed as
\begin{equation}
	\left[ \begin{array}{l}
		{\mathbf{n}_l}\\
		{\mathbf{v}_l}
	\end{array} \right] = \left[ {\begin{array}{*{20}{c}}
			{{{\bf{R}}_l}}&{{{[{{\bf{T}}_l}]}_ \times }{{\bf{R}}_l}}\\
			0&{{{\bf{R}}_l}}
	\end{array}} \right]\left[ \begin{array}{l}
		\mathbf{n}\\
		\mathbf{v}
	\end{array} \right],
\end{equation}
\begin{equation}
	{d_{{l_1}}} = \frac{1}{{\sqrt {{b_x}^2 + {b_y}^2} }}{{\bf{b}}^T}{\bf{K}}_e^l{\mathbf{n}_l},
	\label{eq1-3}
\end{equation}
where ${{\cal L}_l} = {\left( {{{\bf{n}}_l},{{\bf{v}}_l}} \right)^T}$ are the Plücker line coordinate in the left camera frame, and the line coefficients of the projected line $\mathbf{L}_l$ are denoted by ${\left( {{b_x},{b_y},{b_z}} \right)^T}$. In this paper, we employ Plücker coordinates for line triangulation and projection. Nevertheless, it is worth mentioning that these coordinates are overparameterized, being represented by six dimensions, despite the fact that a 3D line inherently possesses only four DOF. During the process of line optimization, additional DOF not only leads to numerical instability but also incurs higher computational costs. Hence, we utilize orthonormal representation $\left(\mathbf{U},\mathbf{W} \right)\in SO\left( 3 \right)\times SO\left( 2 \right)$ of a line $\mathcal{L}$  for optimization~\cite{hartley2003multiple}, which can be derived from the Plücker coordinates using QR decomposition 
\begin{align}
\left[ {{\bf{n}}\left| {{\bf{v}}{\rm{ }}} \right.} \right] = \underbrace {\left[ {\begin{array}{*{20}{c}}
			{\frac{{\bf{n}}}{{\left\| {\bf{n}} \right\|}}}&{\frac{{\bf{v}}}{{\left\| {\bf{v}} \right\|}}}&{\frac{{{\bf{n}} \times {\bf{v}}}}{{\left\| {{\bf{n}} \times {\bf{v}}} \right\|}}}
	\end{array}} \right]}_{\bf{U}}\underbrace {\left[ {\begin{array}{*{20}{c}}
			{\left\| {\bf{n}} \right\|}&0\\
			0&{\left\| {\bf{v}} \right\|}\\
			0&0
	\end{array}} \right]}_\mathbf{\Sigma},
	\label{eq1-5}
\end{align}
where $\mathbf{\Sigma}$ is a diagonal matrix, its two non-zero entries, defined up to scale, can be represented by the matrix $\mathbf{W}$. 

To enhance the precision of the reconstructed 3D lines, we employ a structure-only refinement procedure, which is a special case of bundle adjustment where camera poses are fixed during the optimization process. The cost function that we seek to minimize is composed of the sum of line reprojection errors from the left and right cameras

\begin{equation}
	\underset{\mathcal{L}}{\mathop{\arg \min }}\,\sum\limits_{\left\{ \mathbf{L}_l \right\},\left\{ \mathbf{L}_r\right\}}{\rho \left( {{d}_{l_1}}^{2} + {{d}_{l_2}}^{2} + {{d}_{r_1}}^{2} + {{d}_{r_2}}^{2} \right)},
\end{equation}
where $\rho$ represents the Huber function, which is employed to enhance robustness. ${{d_{{l_1}}},{d_{{l_2}}},{d_{{r_1}}},{d_{{r_2}}}}$ represent the distances from endpoints to the projected lines, as illustrated in Fig.~\ref{fig3}\subref{fig3b}.

\textbf{Endpoint determination.} The above steps aim to optimize the 3D line. However, their precise endpoints have not yet been determined. Endpoint determination is crucial not only for the visualization of the object model, but also for effectively facilitating the subsequent event-line matching. Note that in practice, endpoint uncertainty arises as a result of various factors, including line occlusion and sensor noises. Hence, there may be a lack of strict correspondence between the extracted line endpoints from the left and right cameras. Considering this situation, we employ an effective strategy for endpoint determination by exploiting the characteristics of events. Specifically, this involves comprehensively leveraging the temporal and spatial constraints of stereo event streams to precisely ascertain the corresponding endpoints of lines.

We utilize the spatio-temporal consistency of stereo events to locate the corresponding events, thereby further determining the corresponding endpoints of lines. As illustrated in Fig.~\ref{fig3}\subref{fig3c}, using the extracted lines $\mathcal{L}_{l}$ and $\mathcal{L}_{r}$ as examples, we extract an equal number of equidistant sample events in their vicinity. The event $\mathbf{e}_{i}^{l}=\left\{ \mathbf{p}_{i}^{l},t_{i}^{l} \right\}$ of the left camera is defined by its pixel coordinates ${\mathbf{p}}_{i}^{l}={{\left( x_{i}^{l},y_{i}^{l} \right)}^{T}}$ and the triggered time $t_{i}^{l}$. Similarly, the corresponding event from the right camera can be denoted as $\mathbf{e}_{i}^{r}=\left\{ \mathbf{p}_{i}^{r},t_{i}^{r} \right\}$, where ${\mathbf{p}}_{i}^{r}={{\left( x_{i}^{r},y_{i}^{r} \right)}^{T}}$. Candidate event correspondences are determined by spatio-temporal consistency loss~\cite{ieng2018neuromorphic}, denoted as $\mathbf{C}$, comprising event-epipolar line distances $\mathbf{C}_e$ and temporal distances $\mathbf{C}_t$, both falling below a specified threshold,

\begin{align}
	\mathbf{C} &= \mathbf{C}_e + \mathbf{C}_t,\\
	\mathbf{C}_e &= \frac{d\bigl( \mathbf{p}_i^l, \mathbf{I}_{\mathbf{p}_i^r} \bigr) + d\bigl( \mathbf{p}_i^r, \mathbf{I}_{\mathbf{p}_i^l} \bigr)}{2\epsilon_e},\\
	\mathbf{C}_t &= \frac{\left| t_i^l - t_i^r \right|}{\epsilon_t},
\end{align}
where $d\left( \mathbf{p},\mathbf{I}_{p} \right)$ is the distance between events and the epipolar line $\mathbf{I}_{p}$. ${{\epsilon }_{e}}$ and ${{\epsilon }_{t}}$ are normalizing scalars, representing the maximum permissible geometric distance and the temporal distance, respectively. The correspondence of sampled events can be determined by comparing the loss $\mathbf{C}$. The satisfying event correspondences that are closest to the ends of the line are selected as 2D endpoints. We triangulate the 2D endpoints to derive the endpoints of the 3D line, denoted as ${\mathbf{P}_{i}}=\left[ {\mathbf{X}}_{i},{\mathbf{Y}}_{i},{\mathbf{Z}}_{i},1 \right]^{T}$. According to the pinhole camera model, we have
\begin{equation}
	\mathbf{p}_{i}^{l}=\mathbf{M}_{l}{{\mathbf{P}}_{i}}\text{ },\text{ }\mathbf{p}_{i}^{r}=\mathbf{M}_{r}{{\mathbf{P}}_{i}},
	\label{eq1-11}
\end{equation}
where $\mathbf{M}_{l}$ and $\mathbf{M}_{r}$ represent the projection matrices of the left and right cameras, respectively,
\begin{equation}
	{{\mathbf{M}}_{l}}\!=\!{{\mathbf{K}}_{p}^{l}}\left[ {{\mathbf{R}}_{l}}\left| {{\mathbf{T}}_{l}} \right. \right]\!=\!\left[ \begin{matrix}
		\mathbf{m}_{l_1}^{T}  \\
		\mathbf{m}_{l_2}^{T}  \\
		\mathbf{m}_{l_3}^{T}  \\
	\end{matrix} \right],{{\mathbf{M}}_{r}}\!=\!{{\mathbf{K}}_{p}^{r}}\left[ {{\mathbf{R}}_{r}}\left| {{\mathbf{T}}_{r}} \right. \right]\!=\!\left[ \begin{matrix}
		\mathbf{m}_{r_1}^{T}  \\
		\mathbf{m}_{r_2}^{T}  \\
		\mathbf{m}_{r_3}^{T}  \\
	\end{matrix} \right].
	\label{eq1-12}
\end{equation}

Applying the direct linear transform allows us to recast Eqs.~(\ref{eq1-11}) and~(\ref{eq1-12}) into a homogeneous linear system in Eq.~(\ref{eq1-13}). By performing Singular Value Decomposition on matrix $\mathbf{A}$, we can derive the 3D point ${\mathbf{P}}_{i}$ that minimizes the least square error defined by
\begin{equation}
	\left[ {\begin{array}{*{20}{c}}
			{x_i^l{\bf{m}}_{{l_3}}^T - {\bf{m}}_{{l_1}}^T}\\
			{y_i^l{\bf{m}}_{{l_3}}^T - {\bf{m}}_{{l_2}}^T}\\
			{x_i^r{\bf{m}}_{{r_3}}^T - {\bf{m}}_{{r_1}}^T}\\
			{y_i^r{\bf{m}}_{{r_3}}^T - {\bf{m}}_{{r_2}}^T}
	\end{array}} \right]\left[ {\begin{array}{*{20}{c}}
			{{{\bf{X}}_i}}\\
			{{{\bf{Y}}_i}}\\
			{{{\bf{Z}}_i}}\\
			1
	\end{array}} \right] = {\bf{A}}{{\bf{P}}_i} = 0.
	\label{eq1-13}
\end{equation}

Considering practical scenarios, point ${\mathbf{P}}_{i}$ may not lie exactly on the line $\mathcal{L}$. Therefore, we construct a perpendicular line originating from the point ${\mathbf{P}}_{i}$ and extending towards the line $\mathcal{L}$. Subsequently, we designate the point of intersection between this perpendicular and the line $\mathcal{L}$ as the endpoint of the line $\mathcal{L}$. Once all lines and endpoints are determined, the wireframe model of uncooperative spacecraft can be reconstructed. In close-range measurement scenarios, initialization typically needs to be performed only once, followed by continuous pose tracking. Re-initialization is triggered only when tracking fails.

\begin{figure}[t] 
	\centering{\includegraphics[width=8.5cm]{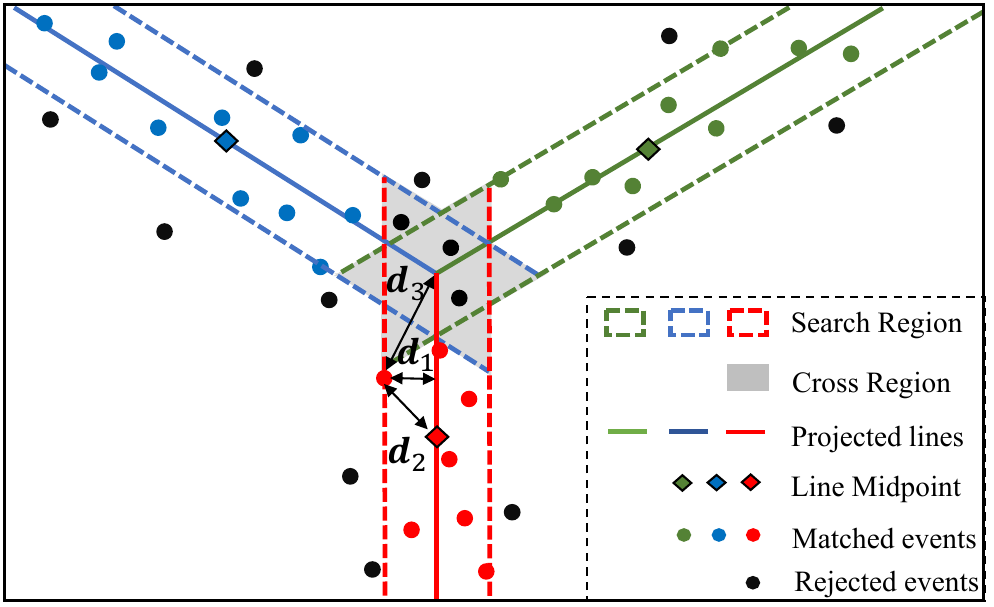}}
	\caption{Illustrative diagram of event-line matching. The matching of events and projection lines is based on distance constraints, where $d_1$ represents the distance from an event to the nearest line segment, $d_2$ represents the distance from an event to the midpoint of the nearest line segment, and $d_3$ represents the distance from an event to the second-nearest line segment.}
	\label{fig4}
\end{figure}

\subsection{Pose tracking}
\label{sec3.2}
In this paper, the tracking module solely relies on stereo event streams without generating image frames. Furthermore, our tracking method directly operates on events, aiming to maximize the utilization of the valuable information contained within them. There is no necessity for us to extract features from event streams in the tracking module, as this may prove to be time-consuming and unreliable. During the initialization stage, the extrinsic parameters of the left camera are set as follows: the rotation matrix ${\bf{R}}_{l}$ is initialized as the identity matrix, and the translation vector ${\bf{T}}_{l}$ is set to zero. On the other hand, the right camera's extrinsic parameters ${\bf{R}}_{r}$ and ${\bf{T}}_{r}$, can be determined by employing the extrinsic parameters of stereo camera calibration ${\bf{R}}_{r2l}$ and ${\bf{T}}_{r2l}$,
\begin{equation}
	{{\bf{R}}_r} = {\bf{R}}_{r2l}^{ - 1}{{\bf{R}}_{l}},{{\bf{T}}_r} = {\bf{R}}_{r2l}^{ - 1}\left( {{{\bf{T}}_l} - {{\bf{T}}_{r2l}}} \right).
\end{equation}

Throughout the maneuvers of uncooperative spacecraft, the stereo event camera is affixed at a location, continuously capturing events. For the newly arrived event cluster, we utilize the pose of the previous event cluster as the initial pose value for the current moment. Once the prior knowledge of the pose is known, the 3D lines of spacecraft are projected, and then the projected lines are associated with events. We conduct a guided search to constrain the complexity and improve the precision of the event-line matching. Subsequently, the current pose of spacecraft is further refined through motion-only bundle adjustment using the Levenberg-Marquardt algorithm.

\textbf{Event-line matching.}
In order to ensure optimal matching between events and projected line segments, we present a straightforward yet effective strategy by iterating through events to locate the closest projected lines. Fig.~\ref{fig4} provides a more intuitive representation of the general concept behind event-line matching. For each event, we calculate its distance to the nearest line segment as $d_1$ and the midpoint of the line segment as $d_2$, to verify that the event is in the vicinity of the line segment. Subsequently, we proceed to calculate the distance from this event to the second nearest line segment as $d_3$, to ensure that the event does not lie within the intersection region of the line segment. If the distance between an event and a line segment satisfies the predefined distance constraints, it is considered a successful match. Otherwise, the event is discarded because it may not be independently triggered by a line. Another advantage of associating events with lines is the effective resolution of occlusion issues, as only visible lines are correlated with events.

\begin{figure*}[tbp]
	\centering
	\subfloat[Rendered experimental scenarios of simulated events.]{
		\begin{minipage}[t]{0.37\textwidth}
			\centering
			\includegraphics[width=\linewidth]{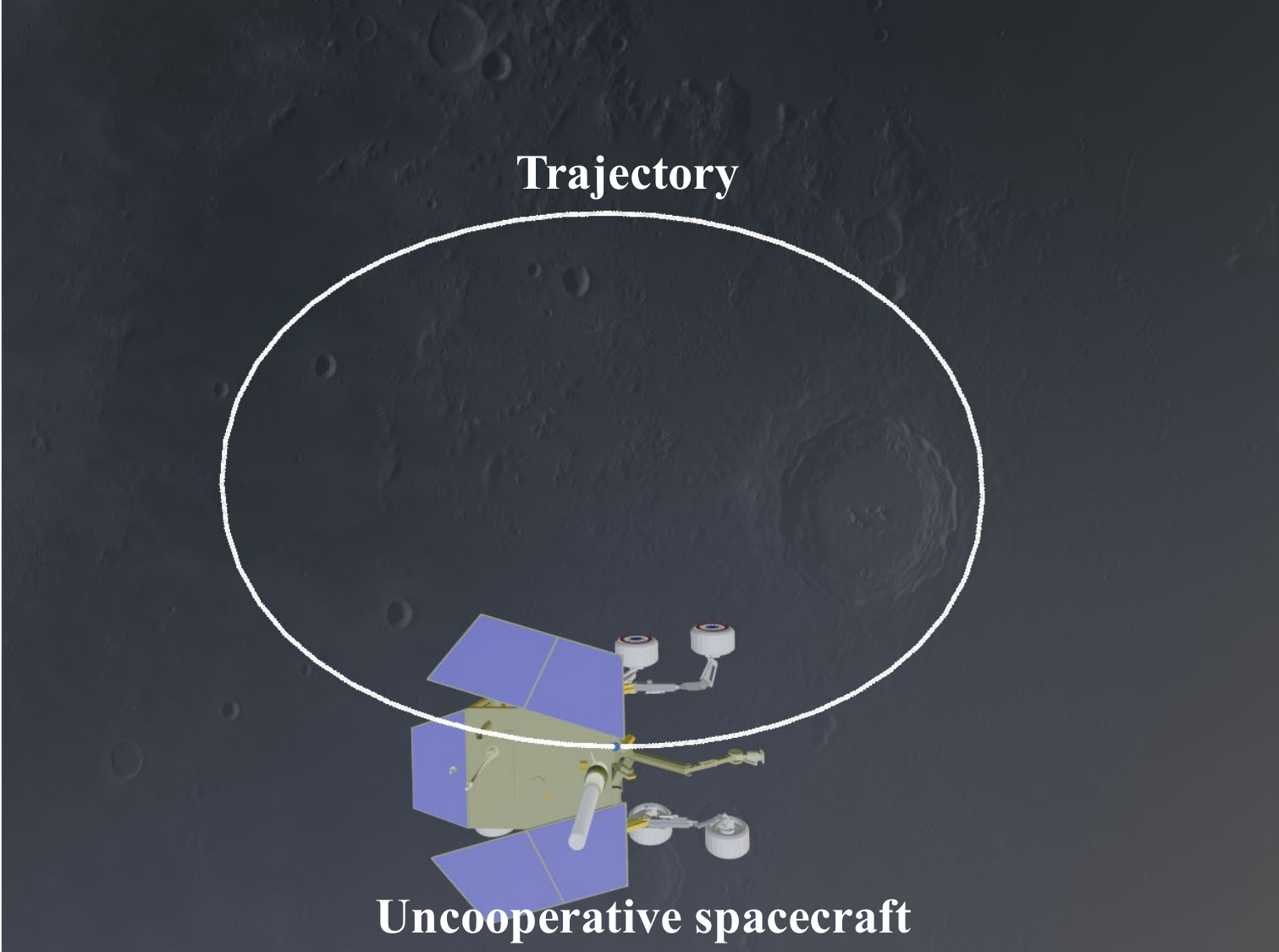}
			\label{fig5a}
		\end{minipage}%
	}
	\subfloat[Experimental scenarios of real events.]{
		\begin{minipage}[t]{0.49\textwidth}
			\centering
			\includegraphics[width=\linewidth]{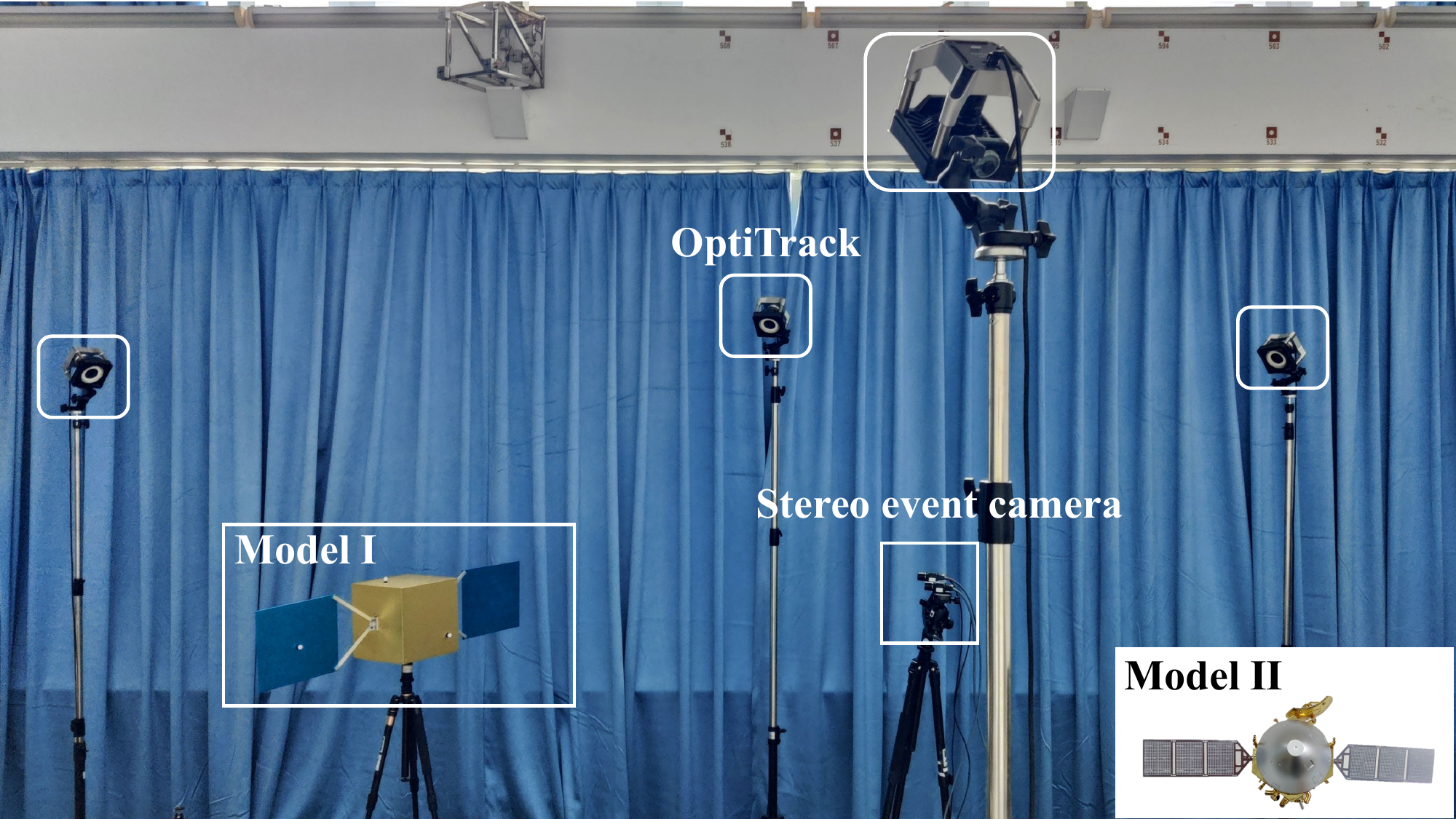}
			\label{fig5b}
		\end{minipage}%
	}
	\caption{Experimental scenarios of our self-collected dataset. (a) Rendered experimental scenarios of Sequence 01. (b) Experimental scenarios of real events. The tested satellite models include Model \uppercase\expandafter{\romannumeral1} and Model \uppercase\expandafter{\romannumeral2}. }
    \label{fig5} 
\end{figure*}

\textbf{Pose optimization.}
We cast the object tracking problem as a joint optimization over the 6-DOF motion parameters using bundle adjustment. For incoming event clusters, we leverage the pose of the previous event cluster as the initial value, and establish associations between events and projected lines via event-line matching. Due to the motion of uncooperative spacecraft, events are triggered, thereby providing the most precise depiction of the current pose of spacecraft. The initial pose is refined by minimizing the distance between events and their corresponding project lines. During this process, events that deviate further from lines are more likely to be outliers or noise, therefore, they are assigned smaller weights in order to mitigate their influence on the pose optimization. Given the event clusters of the left and right cameras $\mathbf{E}_{t}^{l}$, $\mathbf{E}_{t}^{r}$ at time $t$, we optimize the object pose by minimizing event-line distances, using the Huber function

\begin{equation}
	\underset{{{\mathbf{R}}_{l}},{{\mathbf{T}}_{l}}}{\mathop{\arg \min }}\,\sum\limits_{\left\{ {{\mathbf{L}}_{l}} \right\}}{\sum\limits_{\left\{ \mathbf{e}_{j}^{l} \right\}}{\rho \left( {{d}_{le}}^{2} \right)}}+\sum\limits_{\left\{ {{\mathbf{L}}_{r}} \right\}}{\sum\limits_{\left\{ \mathbf{e}_{k}^{r} \right\}}{\rho \left( {{d}_{re}}^{2} \right)}},
\end{equation}

\begin{align}
	{{d}_{le}} &= \frac{1}{\sqrt{{{a}_{lx}}^{2}+{{a}_{ly}}^{2}}}{{\left( \mathbf{e}_{j}^{l} \right)}^{T}}{{\mathbf{K}}_{e}^{l}}\left[ {{\mathbf{R}}_{l}}\left| {{[{{\mathbf{T}}_{l}}]}_{\times }}{{\mathbf{R}}_{l}} \right. \right]{\mathcal{L}},\\
	{{d}_{re}} &= \frac{1}{\sqrt{{{a}_{rx}}^{2}+{{a}_{ry}}^{2}}}{{\left( \mathbf{e}_{k}^{r} \right)}^{T}}{{\mathbf{K}}_{e}^{r}}\left[ {{\mathbf{R}}_{r}}\left| {{[{{\mathbf{T}}_{r}}]}_{\times }}{{\mathbf{R}}_{r}} \right. \right]{\mathcal{L}},
\end{align}
where ${\mathbf{K}}_{e}^{l}$ and ${\mathbf{K}}_{e}^{r}$ represent the intrinsic matrices of the left and right cameras for lines, ${d}_{le}$ and ${d}_{re}$ are the distances from events of the left and right camera to their corresponding projection lines ${\mathbf{L}}_{l}$ and ${\mathbf{L}}_{r}$, respectively. The line coefficients of the projected lines ${\mathbf{L}}_{l}$ and ${\mathbf{L}}_{r}$ are denoted by $\left( {{a}_{lx}},{{a}_{ly}},{{a}_{lz}} \right)$ and $\left( {{a}_{rx}},{{a}_{ry}},{{a}_{rz}} \right)$.

\begin{figure*}[t]
	\centering
	\includegraphics[width=1\linewidth,height=0.5\linewidth]{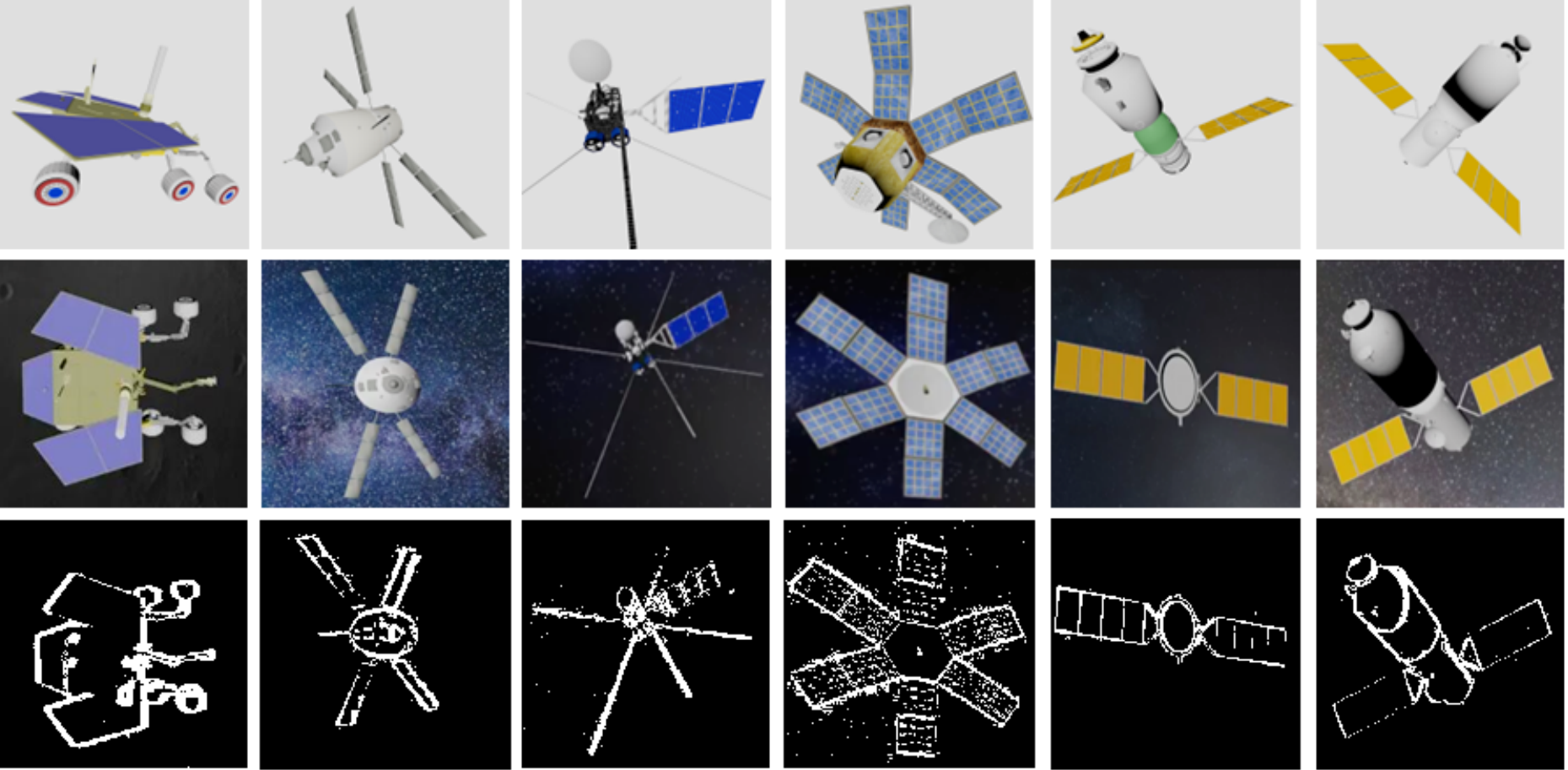}
	\caption{The models, images, and events of uncooperative spacecraft in simulated event datasets. The top row features 3D models of uncooperative spacecraft, the middle row showcases rendered RGB images of uncooperative spacecraft, and the bottom row displays event-accumulated images of uncooperative spacecraft (for visualization purposes only, not utilized). From left to right, they respectively correspond to sequences 01-06.}
	\label{fig6}
\end{figure*}

\section{Experiments}
\label{sec4}
In this section, we evaluate the performance of our pose tracking method for uncooperative spacecraft on simulated and real event datasets. The detailed description of our self-collected datasets is presented in Section~\ref{sec4.1}. Following that, we proceed to conduct simulated and real event experiments in Sections~\ref{sec4.2} and~\ref{sec4.3}. All experiments are conducted on a laptop equipped with a 12th Generation Intel(R) Core(TM) i7-12700H processor with 16GB of RAM.

\subsection{Stereo event-based datasets}
\label{sec4.1}
Owing to the lack of an appropriate object motion dataset captured by stereo event cameras, especially for uncooperative spacecraft, we build a comprehensive dataset to evaluate the performance of our method. Our dataset documents the trajectories of various spacecraft models, encompassing both simulated and real-world events.

\textbf{Simulated event dataset.}
To achieve a more comprehensive simulation of the space environment and the spacecraft motion in orbit, we select six distinct spacecraft models and utilize Blender for photorealistic rendering. Firstly, we determine the parameters of the stereo camera and the trajectories of spacecraft. Subsequently, the spacecraft undergoes continuous 6-DOF motion along its predetermined trajectory with varying velocities, classified as fast, normal, and slow. These spacecraft are assigned different trajectories with different velocities based on their motion characteristics. Furthermore, we assign each spacecraft with distinct spatial backgrounds to comprehensively simulate the complex space environment. The experimental scene of the simulation experiment is depicted in Fig.~\ref{fig5}\subref{fig5a}. The movements of spacecraft are captured by a stationary stereo camera at a frame rate of 30 Hz and a resolution of 640$\times$480 pixels. The spacecraft models and their corresponding rendered RGB images are depicted in the top row and middle row of Fig.~\ref{fig6}. Then, rendered RGB videos are transformed into event streams using V2E~\cite{9523069}, with the relevant parameters set to be consistent with Blender. We accumulate events onto images for visualization, as depicted in the bottom row of Fig.~\ref{fig6}. %It is evident that we have introduced varying degrees of noise into event streams, with the aim of validating the robustness of our method. 
These videos are labeled as sequences 01-06. The duration of spacecraft motion in these sequences spans from 3.5 to 20 seconds. MER (events per second) denotes the mean event rate of the left event camera~\cite{gao2022vector}, which almost ranges from $2\times {{10}^{5}}$ to $1\times {{10}^{6}}$ for these sequences. More experimental details are presented in Table~\ref{table0}. As this method is based on space-based observation systems, both velocity and angular velocity are considered relative quantities. The motion parameters are set to smaller values to ensure long-term observation of spacecraft within the limited camera field of view, aiming to thoroughly validate the tracking method.

\begin{table}[t]
	\centering
	\setlength{\tabcolsep}{4.5pt}
	\begin{tabular}{cccccc}
		\toprule
		Sequence & \thead{EPS\\(event/s)} & \thead{Time \\(s)} & \thead{ Trajectory \\ Length (m)} &  \thead{Angular\\ (deg)}  & \thead{Angular \\Velocity (deg/s)} \\ 
		\midrule
		01-A & 3.43$\times {{10}^{5}}$ & 20.00 & \multirow{2}[1]{*}{11.41}  & \multirow{2}[1]{*}{361.37} & 27.72 \\ 
		01-B & 6.72$\times {{10}^{5}}$ & 10.00 &  & ~  & 55.46 \\
		\midrule
		02-A & 3.10$\times {{10}^{5}}$ & 10.00 & \multirow{2}[1]{*}{10.98}  &\multirow{2}[1]{*}{211.22}  & 55.93 \\ 
		02-B & 6.41$\times {{10}^{5}}$ & 5.00 & ~ & ~  & 111.87 \\
		\midrule
		03-A & 2.34$\times {{10}^{5}}$ & 12.00 & \multirow{2}[1]{*}{6.68} &  \multirow{2}[1]{*}{84.00} &  46.35 \\
		03-B & 4.41$\times {{10}^{5}}$ & 6.00 & ~ & ~ & 92.70 \\
		\midrule
		04-A & 5.47$\times {{10}^{5}}$ & 10.00 & \multirow{2}[1]{*}{8.11} &  \multirow{2}[1]{*}{139.60} & 65.42 \\
		04-B & 1.02$\times {{10}^{6}}$ & 5.00 & ~  & ~ &  32.70 \\ 
		\midrule
		05-A & 3.54$\times {{10}^{5}}$ & 7.00 & \multirow{2}[1]{*}{5.50}  & \multirow{2}[1]{*}{61.52} & 27.58 \\ 
		05-B & 6.76$\times {{10}^{5}}$ & 3.50 & ~ &  ~  & 55.16 \\
		\midrule
		06-A & 5.13$\times {{10}^{5}}$ & 10.00 & \multirow{2}[1]{*}{7.68} &  \multirow{2}[1]{*}{104.32} &  18.04 \\ 
		06-B & 1.09$\times {{10}^{6}}$ & 5.00 & ~ & ~ &  36.07 \\
		\bottomrule
	\end{tabular}
	\caption{The experimental setup of the simulated event datasets.}
	\label{table0}%
\end{table}

\textbf{Real event dataset.}
To evaluate the generality of our models on real-world event-based tracking, we have constructed an experimental platform indoors, as depicted in Fig.~\ref{fig5}\subref{fig5b}. We employ the satellite models as the test object, include Model \uppercase\expandafter{\romannumeral1} and Model \uppercase\expandafter{\romannumeral2}. The installation of markers on the satellite model enables OptiTrack to acquire the ground truth of the satellite's motion trajectory and attitude. During this process, we utilize the pre-calibrated stereo event camera for event acquisition. The stereo rig consists of two Prophesee EVK4 event cameras with a resolution of 1280×720 pixels. We control the satellite to maneuver along distinct trajectories under varying lighting conditions or different velocities, and capture event streams, denoted as sequences 07-11, respectively. Compared to simulated events, actual event cameras are indeed subject to various sources of interference, such as harsh illumination, which can lead to the generation of events with significant noise. The OptiTrack system, which is equipped with actively emitted 850nm near-infrared LEDs, can introduce additional noise due to the flickering effect of these LEDs. To minimize this kind of interference as much as possible, we install light-filtering components in front of the lens. These components are designed to reduce the impact of the flickering effect and improve the signal-to-noise ratio in events captured by the stereo event camera.

\begin{figure*}[tbp]
	\centering
	\subfloat[SGM (RGB image)]{
		\begin{minipage}[t]{0.278\textwidth}
			\centering
			\includegraphics[width=\linewidth]{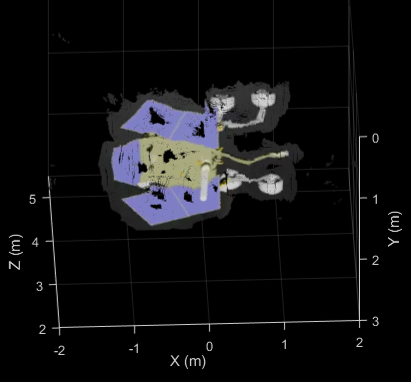}
			\label{fig7a}
		\end{minipage}%
	}
	\subfloat[SGM (Event-accumulated image)]{
		\begin{minipage}[t]{0.28\textwidth}
			\centering
			\includegraphics[width=\linewidth]{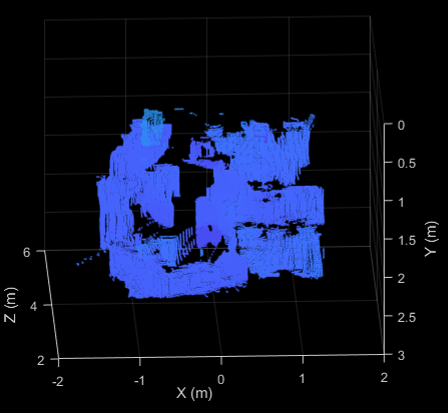}
			\label{fig7b}
		\end{minipage}%
	}
	\subfloat[Our method (Event)]{
		\begin{minipage}[t]{0.346\textwidth}
			\centering
			\includegraphics[width=\linewidth]{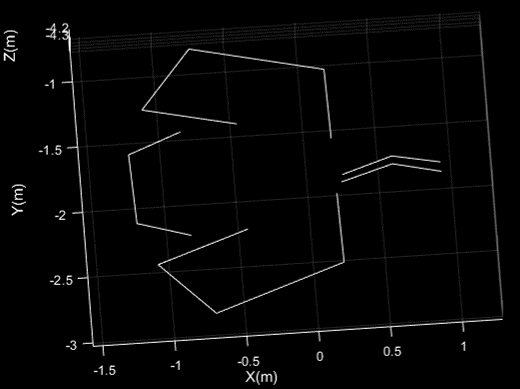}
			\label{fig7c}
		\end{minipage}%
	}
	\caption{Model initialization results for sequence 01. (a) and (b) are the results using SGM, with RGB images and event-accumulated images as inputs, respectively. (c) shows the results of our method, directly applied to events.}
	\label{fig7}
\end{figure*}

\begin{table*}[htbp]
	\centering
	\setlength{\tabcolsep}{7pt}
	\begin{tabular}{cccccccccccccccc}
		\toprule
		Method &\multicolumn{6}{c}{SGM+ICP} &\multicolumn{3}{c}{Line-based}&\multicolumn{3}{c}{ESVO }&\multicolumn{3}{c}{Ours}\\
		\cmidrule(lr){1-1}\cmidrule(lr){2-7}\cmidrule(lr){8-10}\cmidrule(lr){11-13}\cmidrule(lr){14-16}
		Input  &\multicolumn{3}{c}{RGB image} &\multicolumn{3}{c}{Event-accumulated image} &\multicolumn{3}{c}{Event}&\multicolumn{3}{c}{Time-surface map}&\multicolumn{3}{c}{Event}\\
		\cmidrule(lr){1-1}\cmidrule(lr){2-4}\cmidrule(lr){5-7}\cmidrule(lr){8-10}\cmidrule(lr){11-13}\cmidrule(lr){14-16} Sequence&   ${\mathbf{R}_{\text{rel}}}$      &  ${\mathbf{T}_{\text{rel}}}$      & ${\mathbf{T}_{\text{abs}}}$  &     ${\mathbf{R}_{\text{rel}}}$      &  ${\mathbf{T}_{\text{rel}}}$      & ${\mathbf{T}_{\text{abs}}}$ &      ${\mathbf{R}_{\text{rel}}}$    & ${\mathbf{T}_{\text{rel}}}$      & ${\mathbf{T}_{\text{abs}}}$     &  ${\mathbf{R}_{\text{rel}}}$    & ${\mathbf{T}_{\text{rel}}}$     & ${\mathbf{T}_{\text{abs}}}$& ${\mathbf{R}_{\text{rel}}}$    & ${\mathbf{T}_{\text{rel}}}$     & ${\mathbf{T}_{\text{abs}}}$ \\
		\midrule
		01-A & 3.21  & 7.20  & 8.12  &  5.65  & 12.16 & 14.98 & 1.05&3.08&3.58&1.25&2.85&2.13& \textbf{0.97} & \textbf{2.71} & \textbf{2.29} \\
		01-B& 4.51  & 10.54 & 11.59 & 6.46  & 13.30 & 15.74 &1.52&3.70&4.54&1.62&3.24&3.62& \textbf{1.03} & \textbf{2.97} & \textbf{3.07} \\
		\midrule
		02-A & 2.75   & 7.12  & 8.97 &  10.25 & 8.94  & 15.35 &1.55&3.08&3.59&\textbf{1.25}&\textbf{2.85}&2.69&1.37 & 3.44 & \textbf{1.19} \\
		02-B&  3.44 & 8.34  & 11.57 &  12.58 & 13.86 & 14.87  &2.35&3.85&2.73&\textbf{2.04}& 3.75&2.20&2.67  & \textbf{3.70} & \textbf{2.03} \\
		\midrule
		03-A & 10.51 & 25.15 & 24.15 &  12.68 & 27.24 & 50.65 &4.22&14.76&15.69&3.05&11.25&10.68&\textbf{2.14} & \textbf{9.97} & \textbf{9.13}  \\
		03-B& 11.46  & 32.15 & 52.10 &  11.31 & 29.70  & 77.92 & 4.99 &16.81&22.01&4.51&11.02 &12.00&\textbf{2.19} & \textbf{7.04} & \textbf{11.10} \\
		\midrule
		04-A  &5.65  & 7.61  & 9.15 &   5.56  & 10.61 & 16.54 &\textbf{1.27}&\textbf{3.49}&1.54&1.65&3.86&2.15&  1.31& 4.81 & \textbf{1.29}  \\
		04-B& 6.10  & 8.54 & 12.03  & 9.67  & 13.54 & 13.17 & 1.58 &4.93 &2.97&1.85&\textbf{4.02}&2.11& \textbf{1.54}& \textbf{7.42} & \textbf{1.45} \\
		\midrule
		05-A & 2.66  & 5.68  & 8.96 &   2.35  & 7.54  & 13.45 &1.76 &5.28&1.51&2.17&\textbf{4.95}&2.35&  \textbf{1.44} & 5.50 & \textbf{1.21} \\
		05-B& 3.61  & 7.61  & 16.10 &  4.39  & 14.16 & 17.56  & 2.64 &6.91&\textbf{1.53}&3.08&3.75 &2.31&\textbf{2.62} & \textbf{3.69} & 2.26 \\
		\midrule
		06-A & 4.60 &7.31  & 11.04 &  9.46  & 16.79 & 22.46&2.62&3.78& 2.27&2.19&3.38&2.61&  \textbf{2.08}  & \textbf{2.67} & \textbf{2.22}  \\
		06-B &5.33 & 8.95 & 15.25& 13.71 & 8.29  &18.04 &2.71&4.37&\textbf{2.06}&3.04&3.59& 2.94&  \textbf{2.10}  & \textbf{2.68}  & 2.30 \\
		\bottomrule
	\end{tabular}%
	\caption{Comparison of RPE and ATE on simulated event datasets [${\mathbf{R}_{\text{rel}}}$:deg/s, ${\mathbf{T}_{\text{rel}}}$:cm/s, ${\mathbf{T}_{\text{abs}}}$:cm].}
\label{tabel1}%
\end{table*}

\begin{figure*}[t]
	\centering
	\includegraphics[width=1\linewidth,height=0.8\linewidth]{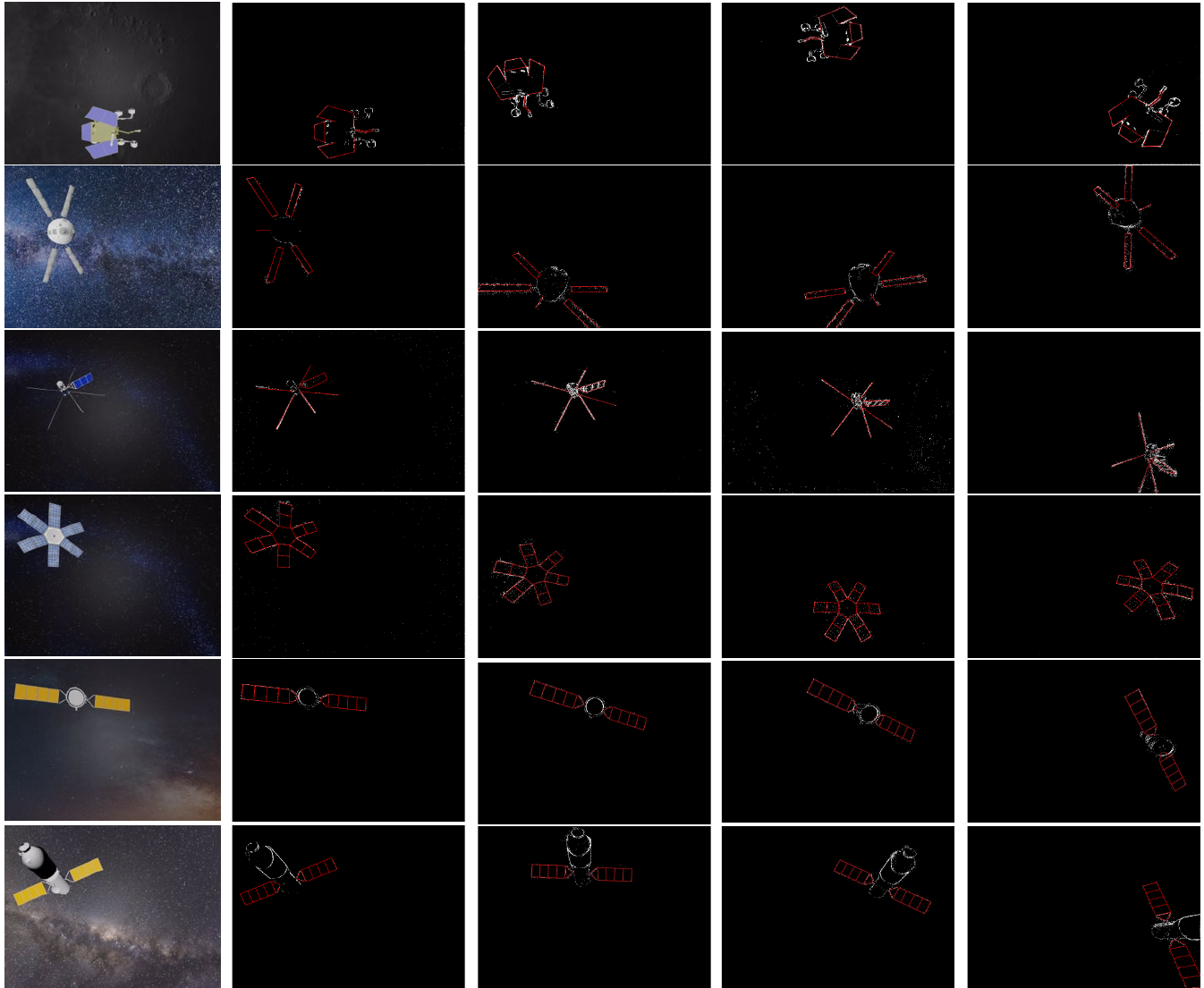}
	\caption{Pose tracking results of uncooperative spacecraft in simulated event experiments. From top to bottom, each row corresponds to events captured by the left camera of sequences 01-06. The first is the rendered RGB image, and on the right is the accumulated image of events (solely for visualization), showcasing the motion of uncooperative spacecraft. Red lines represent the wireframe models of uncooperative spacecraft, which are projected onto the images using the poses solved by our methods.}
	\label{fig8}
\end{figure*}

\subsection{Simulated event experiments}
\label{sec4.2}

Firstly, we extract straight lines directly from event streams of the stereo event camera and reconstruct the wireframe model of spacecraft. To demonstrate the effectiveness of the proposed initialization approach, we conduct a comparative analysis against stereo baseline methods. The baseline method refers to Semi-Global Matching (SGM)~\cite{4359315}, a classic method in the field of stereo vision. RGB images and accumulated event images are separately fed into SGM, and the output results are shown in Fig.~\ref{fig7}\subref{fig7a} and Fig.~\ref{fig7}\subref{fig7b}. The experimental results indicate that SGM performs better on stereo RGB images. However, when it comes to event images that lack texture, color, and other relevant details, the performance of SGM is significantly compromised. The results of the model initialization are illustrated in Fig.~\ref{fig7}\subref{fig7c}. The reconstructed wireframe model is capable of accurately representing the lines of uncooperative spacecraft.

Then, we conduct stereo event-based pose tracking experiments for uncooperative spacecraft. As there are currently no existing stereo event-based object tracking projects available, we compared our method with the following four related and competitive methods.
\begin{itemize}
	\item SGM+ICP: SGM is employed for dense depth estimation and acquiring the 3D point cloud of spacecraft. Iterative Closest Point (ICP)~\cite{121791} is applied to determine the relative pose between the point clouds. 
	\item Line-based: Chamorro et al.~\cite{Chamorroeventline} estimate and track the 6-DoF pose of the event camera, which is an event-based line-SLAM method. For comparison, we adopt the key techniques of the method and apply them for object pose estimation and tracking.
	\item ESVO: ESVO~\cite{9386209} represents the pioneering event-based stereo visual odometry methods, employing a parallel tracking-and-mapping paradigm. We primarily utilize the tracking module of ESVO for spacecraft tracking.
	\item E2VID: E2VID~\cite{8946715} propose a recurrent network to reconstruct videos from event streams. In real event experiments, we employ E2VID to reconstruct grayscale images that are subsequently fed into ``SGM+ICP” for comparative analysis. 
\end{itemize}

In order to quantitatively evaluate the errors of each method, we report the object pose estimation results using two standard metrics~\cite{6385773}: relative pose error (RPE) and absolute trajectory error (ATE). The relative pose error, denoted as ${\mathbf{R}_{\text{rel}}}$ and ${\mathbf{T}_{\text{rel}}}$, is employed to evaluate the disparity between the estimated rotation and translation compared to the ground truth. The ATE ${\mathbf{T}_{\text{abs}}}$ quantifies the differentiation between the estimated trajectory and the ground truth. These metrics provide an assessment of the accuracy and consistency of the estimated poses and trajectories, allowing for a comprehensive evaluation of the performance of these methods.

The ``SGM+ICP" method takes two types of inputs, namely RGB images and event-accumulated images. We conduct tests on both types of inputs. ESVO operates on the time-surface map. Our method works directly on events without estimating intermediate quantities (e.g., reconstructed images), thus simplifying the process. To thoroughly validate the tracking error without introducing additional disturbances, our method, along with the line-based and ESVO approaches, all employ a complete wireframe model for pose tracking. This wireframe model is directly extracted from the object's 3D model, ensuring that our evaluation focuses solely on the tracking performance and not on discrepancies that might arise from model inaccuracies or misalignments.

We set a time interval of 10 ms and select approximately 4,000 nearby events to form an event cluster for the left camera, with the same procedure applied to the right camera. During the tracking process, events can be downsampled to enhance computational speed without compromising the tracking accuracy. To better visualize the pose tracking results, Fig.~\ref{fig8} presents the uncooperative spacecraft tracking results of the simulated event dataset using our methods. The wireframe models are projected using the estimated poses, and annotated with red lines. From top to bottom, each row corresponds to events captured by the left event camera of sequences 01-06.

Table~\ref{tabel1} illustrates the pose tracking errors for sequences 01-06. Compared to slow movements, fast movements tend to exhibit slightly larger pose tracking errors. During the rapid motion of the spacecraft, our method is still capable of achieving stable pose tracking. The experimental results demonstrate that our method surpasses most of the baseline methods in these test sequences. Among them, sequence 03 exhibits the highest overall error primarily due to the spacecraft being situated at a considerable distance from the camera. The limited visibility of spacecraft within the image and the scarcity of generated events contribute to pose errors in both the baselines and our method. The ``SGM+ICP" method exhibits relatively poor reconstruction accuracy, which directly compromises the subsequent tracking precision. Compared to the line-based method, our method leverages stereo event streams, thereby providing more comprehensive observational information and enhancing tracking accuracy. ESVO formulates the problem of pose tracking as edge-map alignment. Differing from this approach, we explicitly establish the correlation between events and lines and refine object poses by minimizing the event-line distance. Table~\ref{tabeltime} summarizes the time consumption of our method on sequence 01. Notably, the line reconstruction process requires a relatively longer duration, while pose tracking is completed more swiftly. Nevertheless, overall, the method is capable of meeting the requirements for relative pose estimation of uncooperative spacecraft in our close-range observation tasks.

\begin{table}[t]
	\centering
	\setlength{\tabcolsep}{15pt}
	\begin{tabular}{lcc}
		\toprule
		Sequence & Line reconstruction & Pose tracking  \\ 
		\midrule
		01-A & 35.36 & 4.76  \\
		01-B & 29.54 & 3.62  \\
		\bottomrule
	\end{tabular}
	\caption{The median time consumption of the proposed method on sequence 01 [unit: ms].}
	\label{tabeltime}%
\end{table}

\begin{figure*}[ht]
	\centering
	\subfloat[The visual representation of pose tracking in challenging scenarios.]{
		\begin{minipage}[t]{0.75\linewidth}
			\includegraphics[width=1\linewidth]{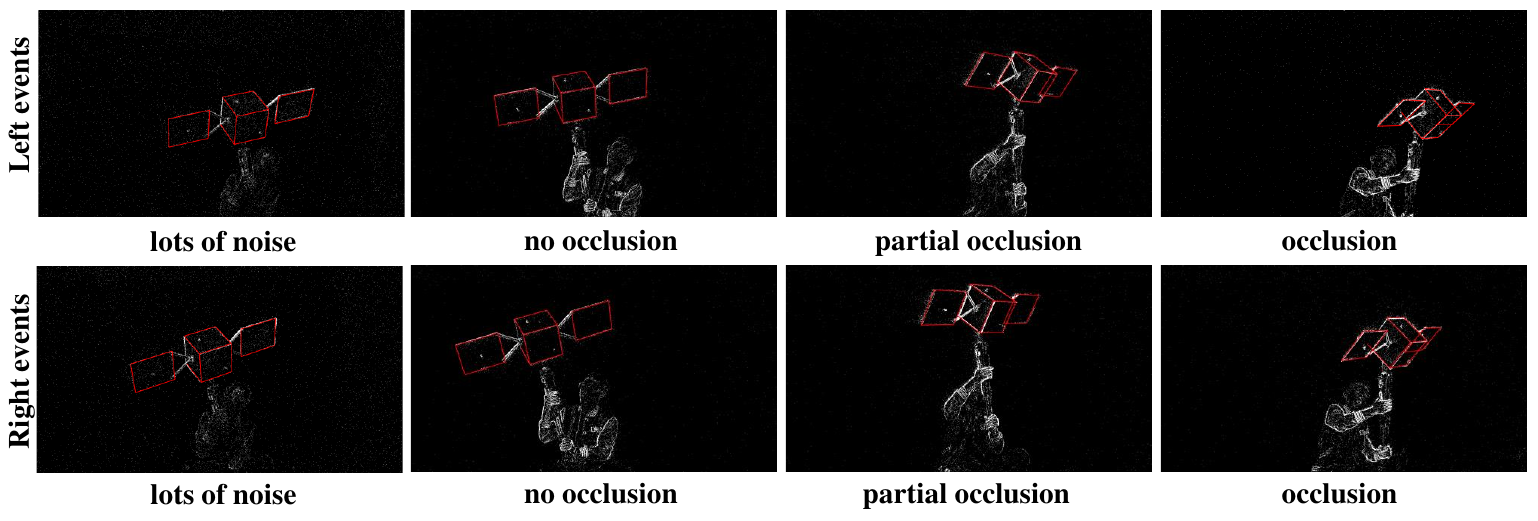}
			\label{fig9a}
		\end{minipage}%
	}
	\subfloat[Trajectory comparisona.]{
		\begin{minipage}[t]{0.25\linewidth}
			\centering
			\includegraphics[width=1\linewidth]{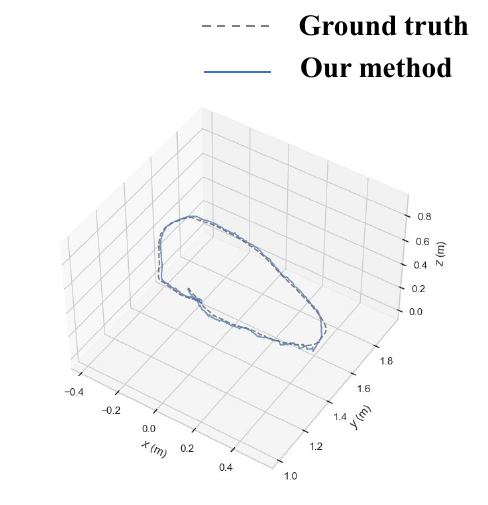}
			\label{fig9b}
		\end{minipage}%
	}
	\caption{Pose tracking results of Model \uppercase\expandafter{\romannumeral1} in sequence 09 of real event experiments. (a) Several challenging scenarios, including noise and occlusion. Red lines represent the reconstructed wireframe model of Model \uppercase\expandafter{\romannumeral1}, which is projected onto event-accumulated images using the poses solved by our method, visually showcasing the pose tracking results. (b) The comparison of the Model \uppercase\expandafter{\romannumeral1} trajectories between our method and the ground truth.}
	\label{fig9}
\end{figure*}

\subsection{Real event experiments}
\label{sec4.3}

We conduct practical experimental verification utilizing self-collected real event datasets. As the EKV4 camera cannot directly capture images, we use E2VID~\cite{8946715} to reconstruct grayscale images and accumulate events into images. Both of these types are input into ``SGM+ICP" for comparison. The proposed method exhibits favorable performance compared to ``SGM+ICP" in practical applications, as evidenced by the outcomes obtained in sequences 07-11. The lack of texture, color, and other information in event-accumulated images, along with event noises, can indeed lead to an unsatisfactory performance of ``SGM+ICP". In contrast, our method is still capable of reconstructing the wireframe model of uncooperative spacecraft followed by continuous pose tracking.

\begin{table*}[htbp]
	\centering
	\setlength{\tabcolsep}{5pt}
	\begin{tabular}{ccccccccccccccccc}
		\toprule
		\multicolumn{2}{c}{Method}  &\multicolumn{6}{c}{SGM+ICP} &\multicolumn{3}{c}{Line-based}&\multicolumn{3}{c}{ESVO }&\multicolumn{3}{c}{Ours}\\
		\cmidrule(lr){1-2}\cmidrule(lr){3-8}\cmidrule(lr){9-11}\cmidrule(lr){12-14}\cmidrule(lr){15-17}
		\multicolumn{2}{c}{Input} &\multicolumn{3}{c}{Grayscale image by ~\cite{8946715}}&\multicolumn{3}{c}{Event-accumulated image} &\multicolumn{3}{c}{Event}&\multicolumn{3}{c}{Time-surface map}&\multicolumn{3}{c}{Event}\\
		\cmidrule(lr){1-2}\cmidrule(lr){3-5}\cmidrule(lr){6-8}\cmidrule(lr){9-11}\cmidrule(lr){12-14}\cmidrule(lr){15-17}  Sequence & Description & ${\mathbf{R}_{\text{rel}}}$      &  ${\mathbf{T}_{\text{rel}}}$      & ${\mathbf{T}_{\text{abs}}}$  &     ${\mathbf{R}_{\text{rel}}}$      &  ${\mathbf{T}_{\text{rel}}}$      & ${\mathbf{T}_{\text{abs}}}$ &      ${\mathbf{R}_{\text{rel}}}$    & ${\mathbf{T}_{\text{rel}}}$      & ${\mathbf{T}_{\text{abs}}}$     &  ${\mathbf{R}_{\text{rel}}}$    & ${\mathbf{T}_{\text{rel}}}$     & ${\mathbf{T}_{\text{abs}}}$& ${\mathbf{R}_{\text{rel}}}$    & ${\mathbf{T}_{\text{rel}}}$     & ${\mathbf{T}_{\text{abs}}}$ \\
		\midrule
		07-Model I&  Low-light & 11.51 & 6.25 & 18.20 &13.54  &8.42  &21.20  &2.04 &1.95&2.35&1.36&1.38&1.64&  \textbf{1.12}& \textbf{0.81} & \textbf{1.48} \\
		08-Model I&  Low-light& 13.65 & 7.41 & 27.59 &15.10&7.25 &23.01 & 1.86&1.25&2.68&1.66&1.08&\textbf{2.54}& \textbf{1.30} & \textbf{0.79} & 2.73 \\
		09-Model I&  Sunlight & 10.28 & 4.54& 24.36 & 9.20&5.22  &16.54  &1.35&0.97&\textbf{2.62}&1.15&1.03 &2.91&\textbf{1.08} & \textbf{0.92} & 3.23 \\
		\midrule
		10-Model II&  Slow speed& 8.45 & 5.76 & 16.53 &11.42&8.65 &15.03 & 1.61&0.58&\textbf{1.29}&1.31&0.88&1.86& \textbf{1.14} & \textbf{0.34} &\textbf{1.29} \\
		11-Model II&  Fast speed & 9.03 & 9.07& 15.74& 12.24& 7.95 &16.01  &2.71&1.68&2.84&\textbf{1.78}&1.16 &2.69&1.92 & \textbf{0.74} &\textbf{2.25}\\
		\bottomrule
	\end{tabular}%
	\caption{Comparison of RPE and ATE on real event datasets [Time:s, MER:event/s, ${\mathbf{R}_{\text{rel}}}$:deg/s, ${\mathbf{T}_{\text{rel}}}$:cm/s, ${\mathbf{T}_{\text{abs}}}$:cm].}
	\label{tabel2}%
\end{table*}

Subsequently, the reconstructed model of objects is utilized for pose optimization and tracking. We set the time interval to 5 ms and select the nearest 20,000 events to generate the event clusters for the left and right cameras, respectively. Similarly, during the tracking process, event downsampling is also performed. For a fair comparison, line-based, ESVO, and our method all employ the same reconstructed model for tracking. By comparing the tracking results with the ground truth obtained from OptiTrack, the relative pose error and absolute trajectory error are shown in Table~\ref{tabel2}. Our method outperforms other approaches on real events, exhibiting smaller errors under different lighting conditions and motion speeds. The pose tracking results of Model \uppercase\expandafter{\romannumeral1} in sequence 09 are shown in Fig.~\ref{fig9}. Red lines represent the reconstructed wireframe model of Model \uppercase\expandafter{\romannumeral1}, which is projected onto event-accumulated images using the poses solved by our method. The top and bottom rows correspond to the pose tracking results of the left and right event cameras, respectively. Fig.~\ref{fig9}\subref{fig9b} illustrates the comparison of the Model I trajectories between our method and the ground truth.

Compared to simulated events, real events incorporate a significant amount of noise, as demonstrated in the first column of Fig.~\ref{fig9}\subref{fig9a}. Undoubtedly, this poses a challenge to the continuous pose tracking of uncooperative spacecraft, resulting in suboptimal performance of the ``SGM+ICP” method. Despite the presence of significant noise in real-world scenarios, our method demonstrates robust tracking capabilities compared to the comparative method. This can be attributed to the effectiveness of our event-line matching strategy, which can effectively distinguish events from noise. Furthermore, during real experimental processes, there are also instances of spacecraft occlusion or partial occlusion, as indicated on the right side of Fig.~\ref{fig9}\subref{fig9a}. The proposed method can effectively address this issue by employing the event-line matching. Edges and lines of spacecraft that are obscured do not generate any events, since we establish associations only between visible lines and events. This association allows for pose optimization by minimizing the distances between events and visible lines. As a result, the pose tracking of uncooperative spacecraft can be continued effectively and accurately. 

\section{Conclusion}
\label{sec5}

This paper proposes a line-based pose tracking method for uncooperative spacecraft with a stereo event camera. First, we reconstruct the wireframe model of uncooperative spacecraft using stereo event streams. Subsequently, our method leverages event-line matching and minimizes event-line distances to achieve pose tracking of uncooperative spacecraft. Experiments have been conducted on our self-collected dataset, which includes multiple sets of simulated events as well as real events. Our method demonstrates superior performance when evaluated on simulated and real events, outperforming competing methods.

\textbf{Future work.} Compared to the tracking module, the model initialization is time-consuming, necessitating further optimization for improved computational efficiency. Furthermore, we plan to extend the methods to handle objects with curved contours, thereby broadening its applicability to a wider variety of uncooperative spacecraft and facilitating robust pose tracking. Additionally, our framework adopts a cluster-based event processing scheme. Inspired by~\cite{glover2024}, it can be extended to support incremental event processing, i.e., operating on an event-by-event basis, enabling asynchronous updates.

\textbf{Limitation.} Our method is primarily designed to handle uncooperative spacecraft with unknown 3D structures. Given that most spacecraft possess geometric structures, such as solar panels, our algorithm leverages these lines for pose estimation. However, for some satellites or damaged spacecraft that lack lines or have curved edge structures, our algorithm may not perform as effectively in such scenarios.

\bibliographystyle{IEEEtran}   
\bibliography{refe}

\end{document}